%% file: main.tex
\documentclass[10pt,twocolumn,letterpaper]{article}

\usepackage{cvpr}              

\usepackage{cvpr}  
\usepackage{multirow}
\input{preamble}
\definecolor{cvprblue}{rgb}{0.21,0.49,0.74}
\usepackage[pagebackref,breaklinks,colorlinks,allcolors=cvprblue]{hyperref}

\def\paperID{10468} 
\def\confName{CVPR}
\def\confYear{2026}

\title{Deeper Thought, Weaker Aim: Understanding and Mitigating Perceptual Impairment during Reasoning in Multimodal Large Language Models}

\author{
Ruiying Peng$^{1}$\thanks{Code: \url{https://github.com/Ivine11/VRGA}}
\quad
Xueyu Wu$^{2}$
\quad
Jing Lei$^{2}$
\quad
Lu Hou$^{2}$
\quad
Yuanzheng Ma$^{1\dagger}$
\quad
Xiaohui Li$^{2\dagger}$ \\
$^{1}$Tsinghua Shenzhen International Graduate School\\
$^{2}$Huawei Technologies\\
{\tt\small pry24@mails.tsinghua.edu.cn}\\
{\small $\dagger$ Corresponding authors}
}

\begin{document}
\maketitle

\input{sec/0_abstract}

\input{sec/1_intro}

\input{sec/2_relatedwork}

\input{sec/3_attention_reasoning}

\input{sec/4_image_attention_patterns}

\input{sec/5_attention_reweighting}

\input{sec/6_experiments_results}

\input{sec/7_conclusions}
\section*{Acknowledgments}
I would like to express my sincere gratitude to all the authors and reviewers for their valuable contributions to this research. This work was supported by Tsinghua Shenzhen International Graduate School and Huawei 2012 Lab.
{
    \small
    \bibliographystyle{ieeenat_fullname}
    \bibliography{main}
}


\end{document}

%% file: sec/0_abstract.tex
\begin{abstract}

Multimodal large language models (MLLMs) often suffer from perceptual impairments under extended reasoning modes, particularly in visual question answering (VQA) tasks. We identify attention dispersion as the underlying cause: during multi-step reasoning, model's visual attention becomes scattered and drifts away from question-relevant regions, effectively “losing focus” on the visual input. To better understand this phenomenon, we analyze the attention maps of MLLMs and observe that reasoning prompts significantly reduce attention to regions critical for answering the question. We further find a strong correlation between model’s overall attention on image tokens and the spatial dispersiveness of model’s attention within the image.~Leveraging this insight, we propose a training-free \textbf{V}isual \textbf{R}egion-\textbf{G}uided \textbf{A}ttention \textbf{(VRGA)} framework that selects visual heads based on an entropy–focus criterion and reweights their attention, effectively guiding the model to focus on question-relevant regions during reasoning.  
Extensive experiments on vision-language benchmarks demonstrate that our method effectively alleviates perceptual degradation, leading to improvements in visual grounding and reasoning accuracy, while offering interpretable insights into how MLLMs process visual information.
\end{abstract}

%% file: sec/1_intro.tex
\section{Introduction}

\begin{figure}[htbp]
    \centering
    \includegraphics[width=1.0\linewidth]{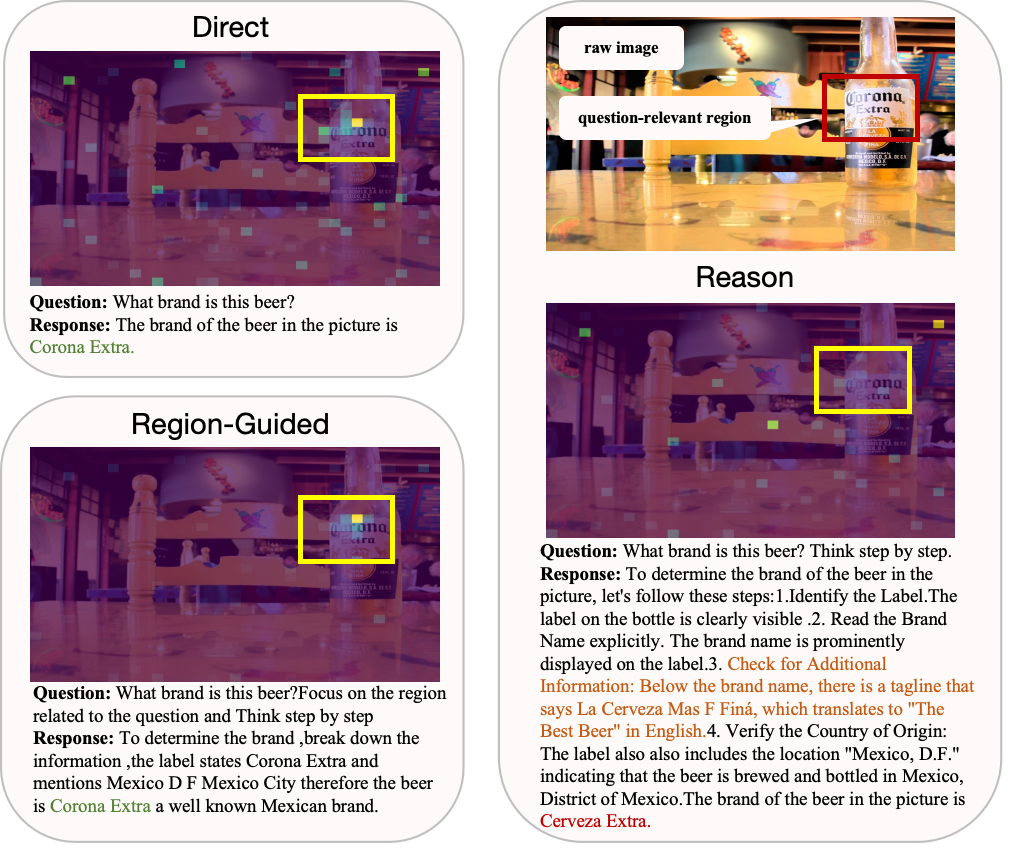}

\caption{\textbf{Comparison of Prompting Strategies in VQA Tasks.}
This figure compares three prompting strategies—\textit{Direct}, \textit{Region-Guided}, and \textit{Reason}—in visual question answering (VQA) tasks.
The attention maps show that the \textit{Direct} mode focuses correctly on question-relevant regions, while the \textit{Region-Guided} approach further reduces attention to irrelevant areas, enhancing visual grounding.
In contrast, the \textit{Reason} mode disperses attention across the scene and often guides the model to describe question-irrelevant regions, leading to incorrect or misleading summaries.
In the QA results, green boxes highlight correct answers, whereas red boxes indicate errors and irrelevant reasoning.
Overall, these visualizations illustrate how different prompting strategies influence attention allocation and task performance.}

    \label{fig:vqacase}
\end{figure}

\begin{figure}[htbp]  
    \centering
    \includegraphics[width=1\linewidth]{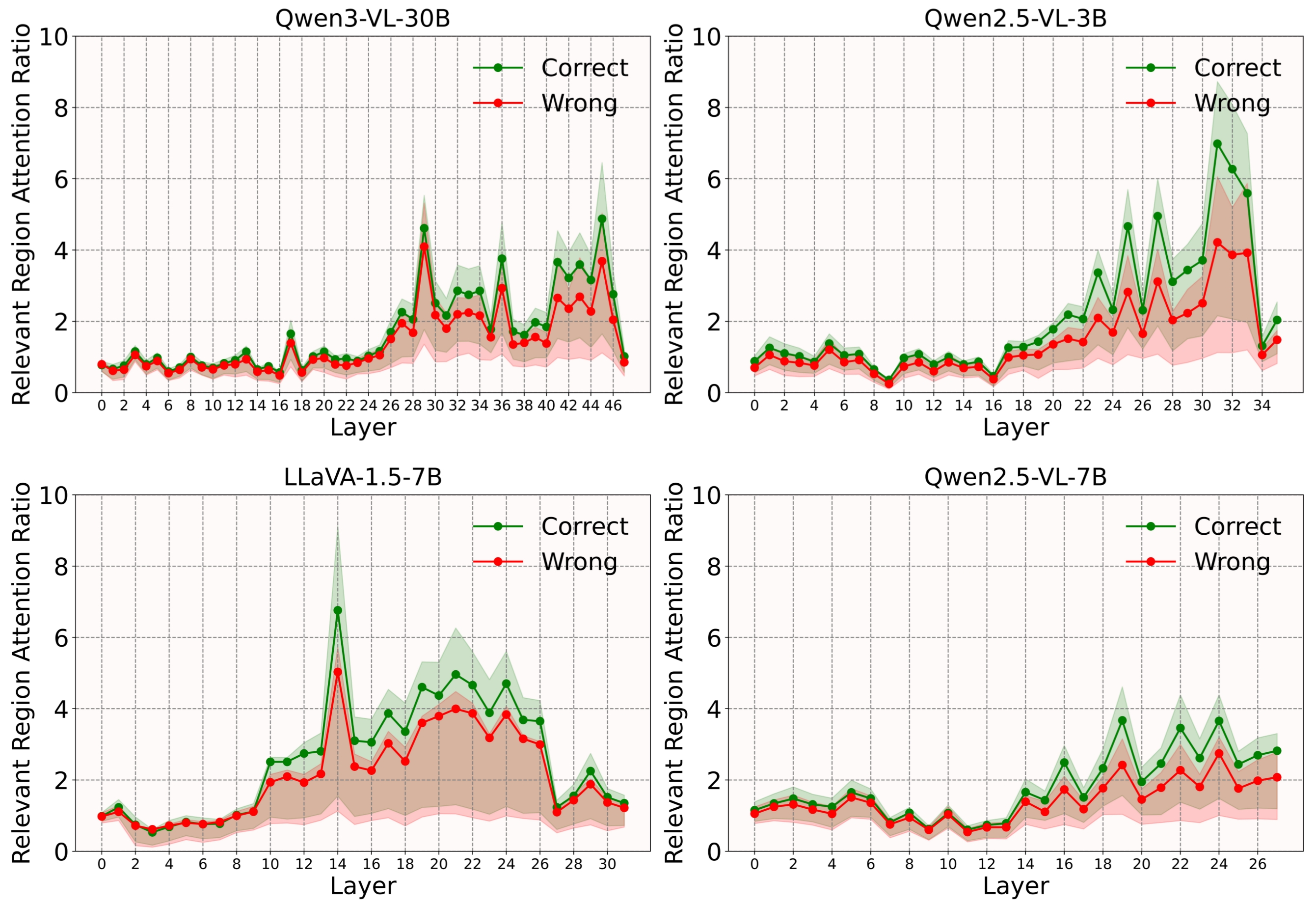}
    \caption{
\textbf{Layer-wise relevant region attention ratio across models.}
The vertical axis denotes the \textit{Relevant Region Attention Ratio}, which measures the degree of attention allocated to question-relevant regions during VQA (as defined in Sec.~\ref{sec:RRAR}).
The horizontal axis represents the Transformer layer index, where the attention maps are averaged across heads.
Green and red curves correspond to correct and wrong predictions, respectively.
The solid line shows the mean value, and shaded areas denote ranges.
Across all models, correct answers consistently exhibit higher attention to relevant regions, suggesting that focusing on the correct visual evidence is crucial for accurate reasoning.
}

    \label{fig:image_attention_T_F}
\end{figure}

Researchers have long aimed to enable LLMs to perform more complex tasks that require deeper reasoning. In the text domain, deeper reasoning has shown progress on mathematical and scientific problems~\cite{math,lu2025scp,yuan2025gsm8k,ling2017program}. By constructing reasoning datasets for supervised fine-tuning (SFT) and using rule-based rewards to encourage correct answers and longer reasoning chains, researchers have trained MLLMs~\cite{Qwen25VL,li2022blip,LLaVA,gemini,Qwen2VL,Qwen-VL,qwen3technicalreport} to adopt similar reasoning paradigms, aiming to improve visual reasoning tasks~\cite{wang2024measuring,lu2022learn,guan2024hallusionbench,wang2024haloquest}.


\textbf{However, a paradox emerges}: recent studies~\cite{yang2025thinking,liu2025more,liao2025improved} reveal that directly generating extended CoT reasoning in VQA can actually \textit{reduce} accuracy compared to direct answering, especially on \textbf{perceptually grounded} tasks~\cite{wang2024haloquest,mmhal,mmstar,sqa}.  Previous works \cite{icot, deepeyes} attribute this degradation to the drop of perception capabilities during reasoning process, \textit{i.e., perceptual accuracy degrades within the reasoning chain}. Thus, they attempt to mitigate it by enhancing perceptual correctness through (1) tool-based extraction of image regions, and (2) injecting image tokens corresponding to reasoning steps into the generation process. However, our analysis on \textbf{TextVQA}~\cite{textvqa} reveals that a majority of the failure cases contain predominantly correct perceptual descriptions (see orange-marked examples in Fig. ~\ref{fig:vqacase}), yet still produce incorrect final answers. This indicates that perceptual capability in the reasoning process is not as weakened as expected. The degradation appears to stem not from \textit{what} visual information is available, but from \textit{how} the model processes it.

\textbf{What causes this reasoning degradation then?} Liu et al.~\cite{liu2025more} found that longer reasoning chains reduce overall attention to images, amplifying visual hallucinations. However, they mainly analyze the correlation between reasoning length and hallucination, while dynamically controlling reasoning length in practice remains difficult, leaving the underlying mechanism largely unexplained. 
Building on this insight, we conduct a comprehensive attention analysis across \textbf{models}~\cite{ming2025ocean,thinklite,mmeureka,LLaVA,Qwen25VL} and uncover a more specific mechanism: common CoT prompts (e.g., ``think step by step") cause attention to become \textit{spatially dispersed}, reducing focus on question-relevant regions—those image areas containing information necessary to answer the question (Fig.~\ref{fig:vqacase}). 
Combined with the prevalence of irrelevant visual descriptions in failure cases, \textbf{we identify the core problem that leads to reasoning degradation: MLLMs ``lose focus" during visual reasoning, attending to task-irrelevant image regions.}

To validate this hypothesis, we first analyze layer-wise attention specifically on question-relevant regions, annotated with ground-truth bounding boxes in the TextVQA dataset. As shown in Fig.~\ref{fig:image_attention_T_F}, correctly answered questions maintain  a higher attention ratio on these regions compared to incorrect predictions across different models.  This indicates the high importance of ``focusing on relevant regions" during reasoning.  Furthermore, with a more comprehensive analysis across multiple models (Fig.~\ref{fig:AttAllocationCombination}) on the attention patterns difference between their reasoning mode and direct mode, we find that: on reasoning mode, the  attention of the models are shifted to irrelevant regions.  To quantify this behavior at inference time, we further introduce more detailed attention patterns analysis and unveil a novel finding: heads that genuinely process visual information exhibit both high attention to image tokens ($R_\text{img}$) and spatially concentrated attention patterns (low entropy  $H_\text{img}$).  A strong linear correlation between these two quantities is  shown in Fig.~\ref{fig:SelectHeadsResult}.  This is because effective visual processing requires not just looking at the image, but focusing on specific regions, unlike text-processing heads that distribute attention broadly across token sequences.    This strong positive correlation supports our hypothesis, and motivates another research question: \textit{Can we restore perceptual grounding and improve reasoning accuracy by selectively enhancing attention to question-relevant areas?}

The key challenge lies in identifying (1) which attention heads are responsible for visual processing, and (2) which image regions are relevant to the question—both at inference time without ground truth. Existing methods ~\cite{zhang2025mllms} perform layer-level selection based on overall image attention, but we observe that heads within the same layer contribute unequally to visual processing. We address both challenges through a \textbf{training-free dynamic approach VRGA}: First, we exploit our key finding -- the correlation between $R_\text{img}$ and  $H_\text{img}$ to automatically identify visual heads without training.  Second, we identify question-relevant regions at inference time by using the question tokens cross-attention to images. We then selectively enhance attention in the identified visual heads specifically to these regions during the reasoning process.

Across all evaluated benchmarks, VRGA delivers stable improvements in comprehensive scores (1–6 points across model scales). Unlike training-based approaches  \cite{icot, deepeyes} that inject additional tokens globally, our method provides \textit{selective, context-aware} attention reweighting only where needed, preserving the model's reasoning fluency while restoring visual grounding.


In summary, our contributions are:
\begin{itemize}
    \item We reveal that CoT prompts cause MLLMs to lose focus on question-relevant regions through attention dispersion, providing a mechanistic explanation for CoT-induced performance degradation in VQA. 
    \item We show that high attention to the entire image does not imply correct visual processing. Only attention heads with concentrated, spatially focused patterns consistently attend to question-relevant regions, and their processing of visual information is strongly correlated with correct answers.
    \item We propose a Visual Region-guided Attention (VRGA) framework that first identifies question-relevant visual tokens via adaptive head selection and attention refinement, and then reweights attention during response generation. This approach effectively guides multimodal LLMs to focus on the correct visual regions, which restores perceptual grounding and improves reasoning accuracy.
\end{itemize}


%% file: sec/2_relatedwork.tex
\begin{figure*}[htbp]  
    \centering
    \includegraphics[width=1\linewidth]{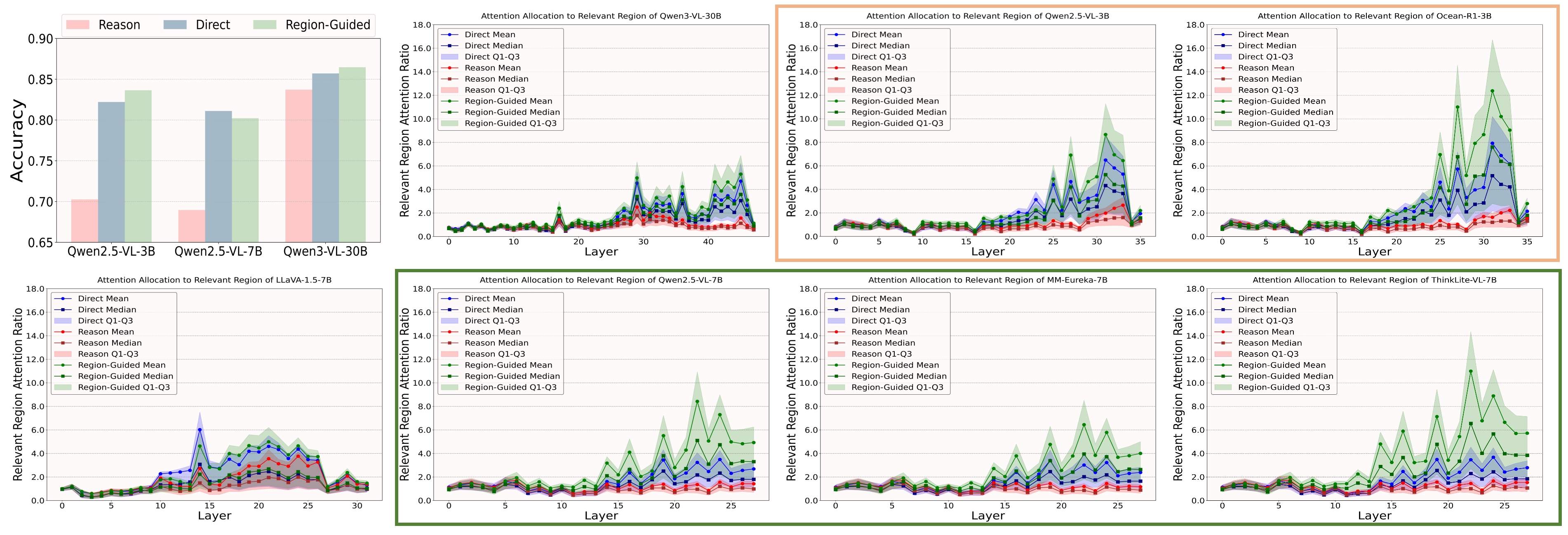}
\caption{
    \textbf{Impact of prompting strategies on visual grounding and performance.}
    The bar chart (left) compares TextVQA accuracy under three prompting strategies: \textit{Reason}, \textit{Direct}, and \textit{Region-Guided}.
    The line plots show the layer-wise RRAR (as defined in Sec.~\ref{sec:RRAR}) from the question-end token to relevant visual regions, which is used to measure the focus on question-related areas.
    Shaded regions indicate the interquartile range (Q1--Q3) of RRAR across samples.
    The \textit{Ocean-R1-3B}~\cite{ming2025ocean} model is built on the \textit{Qwen2.5-VL-3B} backbone, while \textit{MM-Eureka-7B}~\cite{mmeureka} and \textit{ThinkLite-VL-7B}~\cite{thinklite} are based on the \textit{Qwen2.5-VL-7B} backbone.
    For detailed explanation of the relevant region attention ratio (RRAR), please refer to Section~3.
    We observe that the \textit{Reason} mode, which emphasizes sequential reasoning, tends to disperse visual attention and reduce focus on question-relevant regions, leading to lower accuracy.
    By contrast, the proposed \textit{Region-Guided} prompting re-concentrates attention toward relevant regions, achieving both improved grounding and higher task accuracy.
}
    \label{fig:AttAllocationCombination}
\end{figure*}more

\section{Related Work}

\begin{itemize}
\item{\textbf{Enhancing Visual Perception during Reasoning}: To mitigate CoT degradation, several approaches enhance visual-textual integration during reasoning.  ICoT \cite{icot} interleaves image patches at each reasoning step to maintain continuous visual grounding. DeepEyes \cite{deepeyes} introduces visual evidence tokens trained with supervised signals to reference specific regions. CCoT  \cite{ccot} integrates detected objects and spatial relationships as structured reasoning components, while DDoT \cite{chang2025ddot} separates perception and reasoning into dedicated modules.  These methods require additional training on annotated reasoning datasets \cite{icot,deepeyes}, inject visual information globally (potentially introducing irrelevant details), and focus on \textit{what} visual information to include rather than \textit{where} to attend. Our analysis shows that even with all relevant visual information available, CoT prompts cause spatially dispersed attention, reducing focus on question-relevant regions—an attention allocation problem unaddressed by prior work.}

\item{\textbf{Attention Analysis in Vision-Language Reasoning Models}: Recent work analyzes attention mechanisms to understand MLLM reasoning. Zhang et al.~\cite{zhang2025mllms} identify layers responsible for visual processing using overall attention ratios, enabling layer-level optimization. Liu et al.~\cite{liu2025more} show that reduced image attention correlates with hallucinations, but analyze \textit{overall} attention rather than its \textit{spatial distribution}. These studies operate at layer-level granularity and do not distinguish attention heads within layers or differentiate \textit{focused} (spatially concentrated) from \textit{dispersed} attention. Li et al.~\cite{kang2025see} further identify \textit{attention sinks} and exclude such heads from visual-processing heads, but sink-like patterns may still occur on question-relevant regions, leading to misclassification. Moreover, existing analyses remain primarily diagnostic rather than interventional.}


\end{itemize}

We differ from prior work in two aspects. First, we extend attention analysis from coarse layer-level to fine-grained head-level, revealing that visual heads exhibit high image attention ($R_\text{img}$) and low spatial entropy ($H_\text{img}$), indicating spatially concentrated processing. Second, we move from diagnosis to intervention: leveraging the strong correlation between $R_\text{img}$ and $H_\text{img}$, we automatically identify visual heads and selectively enhance their attention to question-relevant regions during reasoning. This training-free approach mitigates attention dispersion at the mechanism level without additional training or global input modification.

%% file: sec/3_attention_reasoning.tex
\section{Understanding Visual Attention in Multimodal Reasoning}

Chain-of-thought prompting disperses visual attention and degrades VQA performance, but the underlying mechanism remains unclear. 
In this section, we analyze how MLLMs process visual information during reasoning.
Prior work~\cite{liu2025more,zhang2025mllms} characterizes visual processing by measuring overall attention to image tokens—the total attention weight assigned to all visual tokens. However, this coarse-grained metric cannot distinguish between \textit{focused} attention (concentrated on specific relevant regions) and \textit{dispersed} attention (spread uniformly across the image).

Our analysis of TextVQA failures indicates that models often describe scenes accurately but still answer incorrectly, often focusing on irrelevant regions (see Fig.~\ref{fig:vqacase}). This suggests that \textbf{perceptual accuracy alone does not guarantee correct reasoning}—models must not only perceive visual content accurately but also \textit{focus} on question-relevant regions while filtering out distractions.

We formalize this intuition through two research questions:
\begin{itemize}
    \item Does CoT prompting systematically disperse attention away from question-relevant regions, and does this dispersion correlate with answer accuracy?
    \item At the head level, do attention heads with stronger visual grounding also exhibit more spatially concentrated attention patterns?
\end{itemize}

\subsection{Relevant Region Attention Ratio}
\label{sec:RRAR}
To measure attention focus on question-relevant regions, we introduce the \textbf{Relevant Region Attention Ratio (RRAR)}, which quantifies the relative attention allocated to question-relevant regions compared to the entire image.

Let $\mathcal{V}$ denote the set of all visual tokens, and $\mathcal{B} \subseteq \mathcal{V}$ the subset corresponding to question-relevant regions identified via ground-truth bounding boxes. Given the attention tensor $\mathbf{A}_{\text{qt}} \in \mathbb{R}^{L \times H \times |\mathcal{V}|}$ representing attention from the final question token to visual tokens across $L$ layers and $H$ heads, we define RRAR for a single head $(l,h)$ as:

\begin{equation}
\Gamma^{(l,h)} = 
\frac{\frac{1}{|\mathcal{B}|}\sum_{i \in \mathcal{B}} a_i^{(l,h)}}
{\frac{1}{|\mathcal{V}|}\sum_{j \in \mathcal{V}} a_j^{(l,h)}},
\quad 
a_i^{(l,h)} \in \mathbf{A}_{\text{qt}}^{(l,h)}
\end{equation}

Intuitively, $\Gamma^{(l,h)} > 1$ indicates the head attends more to relevant regions than the image average, while $\Gamma^{(l,h)} < 1$ indicates dispersed or unfocused attention. For layer-level analysis, we compute the average RRAR across all heads:

\[
\overline{\Gamma}^{(l)} = \frac{1}{H}\sum_{h=1}^{H} \Gamma^{(l,h)}.
\]

\subsection{CoT Prompts Disperse Attention and Reduce Accuracy}


We evaluate seven models (e.g., Qwen2-VL, LLaVA-1.5) on 1,450 TextVQA~\cite{textvqa} samples under three prompting strategies: \textit{Direct}, \textit{Reason}, and \textit{Region-guided} (as shown in Fig.~\ref{fig:vqacase}).

\textbf{Finding 1: Correct answers receive higher RRAR.}  
As shown in Fig.~\ref{fig:image_attention_T_F}, across all models and layers, questions answered correctly exhibit significantly higher RRAR than incorrect answers. This establishes a strong positive correlation between attention focus and reasoning accuracy.

\textbf{Finding 2: CoT prompts systematically reduce RRAR.}  
Fig.~\ref{fig:AttAllocationCombination} shows that reasoning prompts yield substantially lower layer-averaged RRAR than direct answering:

\begin{equation}
\overline{\Gamma}_{\text{reason}} < \overline{\Gamma}_{\text{direct}} < \overline{\Gamma}_{\text{region-guided}}
\end{equation}
Here, $\overline{\Gamma}_{\text{reason}}$, $\overline{\Gamma}_{\text{direct}}$, and $\overline{\Gamma}_{\text{region-guided}}$ denote the RRAR computed under \textit{Reason}, \textit{Direct}, and \textit{Region-guided} prompts, respectively.
This confirms that step-by-step reasoning disperses attention away from question-relevant regions.

\textbf{Finding 3: Attention dispersion correlates with accuracy degradation.}  
Fig.~\ref{fig:AttAllocationCombination} (upper left) shows that direct answering consistently achieves higher accuracy than CoT reasoning. Notably, region-guided prompts improve accuracy, demonstrating that guiding attention toward relevant regions mitigates dispersion.

\textbf{Implication:}  
These findings validate our hypothesis that CoT prompts cause MLLMs to lose focus on question-relevant regions, and this attention dispersion directly impairs reasoning accuracy. This motivates our approach of \textit{directly reweighting attention distributions} to restore focus during reasoning, rather than modifying inputs or training new models.

%% file: sec/4_image_attention_patterns.tex


\begin{figure}[ht]
    \centering
    \includegraphics[width=1\linewidth]{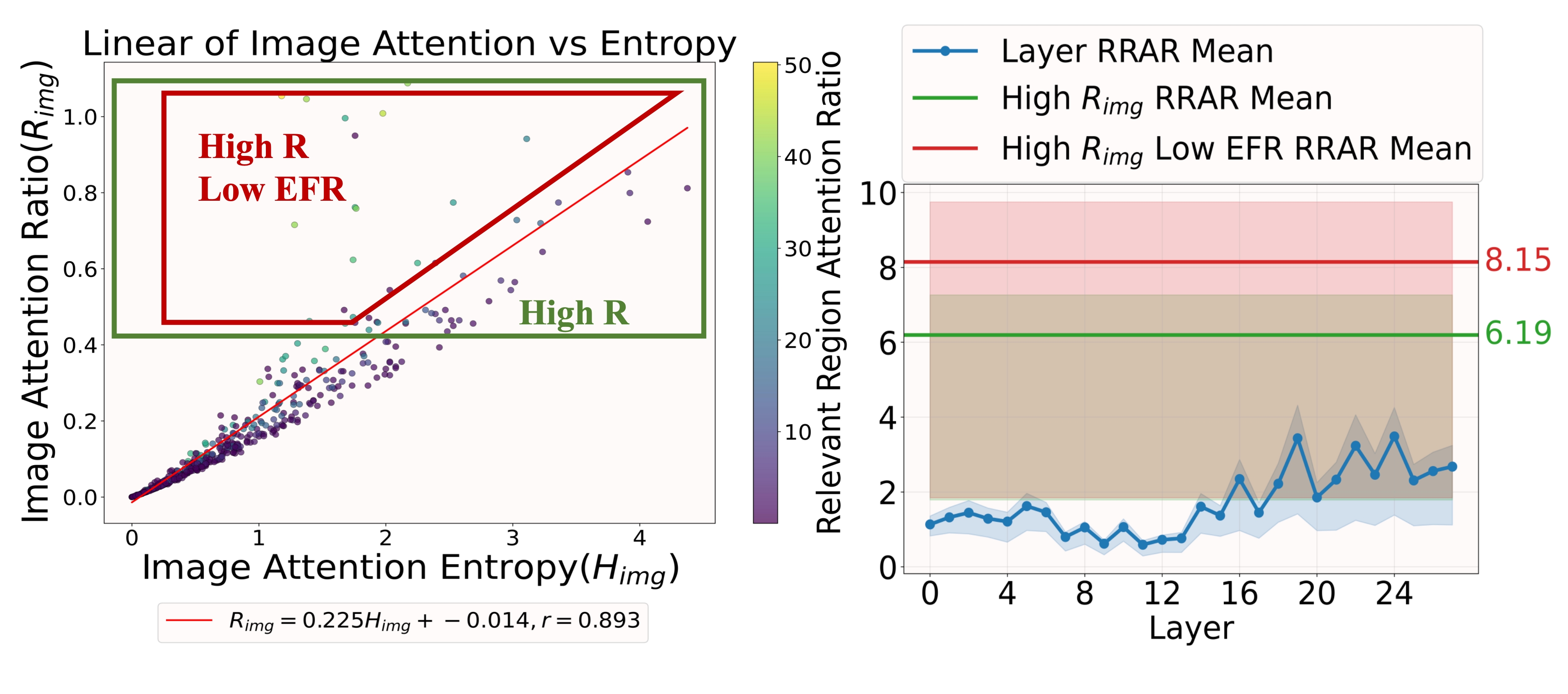}
    \caption{Attention mechanism analysis in vision models. 
    \textbf{Left}: Scatter plot showing the relationship between $R_{img}$ and $H_{img}$. Points within the red box represent heads indicating effective processing of visual information.A linear fit is performed on the points, with the correlation coefficient $r$ representing the Pearson correlation. \textbf{Right}: Line graph comparing the Relevant Region Attention Ratio (RRAR) for different selection methods. The red line represents the RRAR for heads selected by our method, which considers both high $R_{img}$ and low EFR($\frac{H_{img}}{R_{img}}$). The green line shows the RRAR for heads based solely on $R_{img}$. The blue line indicates the RRAR across layers. \textbf{Our method consistently identifies heads that precisely focus on relevant visual regions.}}
    \label{fig:SelectHeadsResult}
\end{figure}

\subsection{Head-Level Analysis: Visual Processing Requires Focused Attention}
\label{sec:visualHead}

The previous analysis establishes that attention dispersion impairs reasoning accuracy, but operates at the layer level by averaging across all heads. However, attention heads within the same layer can serve fundamentally different functions---some may process visual information while others handle linguistic or cross-modal reasoning. To understand the mechanism of visual attention at a finer granularity, we perform head-level analysis to identify which specific heads are responsible for visual processing.


\textbf{Metrics for Head-Level Analysis:} To characterize attention patterns at the head level without requiring ground-truth region annotations, we introduce two complementary metrics:

\begin{itemize}
    \item \textbf{Image Attention Ratio ($R_\text{img}$)} measures \textit{how much} attention a head allocates to visual tokens:
    \begin{equation}
    R_\text{img}^{(l,h)} = 
    \frac{\frac{1}{|\mathcal{V}|}\sum_{i \in \mathcal{V}} a_i^{(l,h)}}
         {\frac{1}{M}\sum_{j=1}^{M} a_j^{(l,h)}}, 
    \end{equation}
    where $\mathcal{V}$ denotes visual tokens with $|\mathcal{V}| = N$, $M$ is the total number of tokens (visual + text), and $a_i^{(l,h)} \in \mathbf{A}_{\text{qt}}^{(l,h)}$ is the attention weight from the final question token to token $i$ at layer $l$, head $h$. 
    
    \item \textbf{Image Attention Entropy ($H_\text{img}$)} measures \textit{how concentrated} attention is within visual tokens:
    \begin{equation}
    \tilde{a}_i^{(l,h)} = \frac{a_i^{(l,h)}}{\sum_{j \in \mathcal{V}} a_j^{(l,h)}}, 
    \end{equation}
    \begin{equation}
    H_\text{img}^{(l,h)} = - \sum_{i \in \mathcal{V}} \tilde{a}_i^{(l,h)} \log(\tilde{a}_i^{(l,h)} + \epsilon)
    \end{equation}
    where $\epsilon = 10^{-8}$ prevents numerical instability. Lower $H_\text{img}$ indicates spatially concentrated (focused) attention, while higher $H_\text{img}$ indicates spatially dispersed attention across the image.
\end{itemize}

Together, $R_\text{img}$ and $H_\text{img}$ characterize both the \textit{amount} and \textit{focus} of visual attention, enabling us to distinguish between heads that merely attend to images versus heads that focus effectively on relevant regions.

\textbf{Visual Heads Exhibit High Attention and Low Entropy:} We analyze attention heads across 5 models on 1500 samples from TextVQA. For each head, we compute $R_\text{img}$, $H_\text{img}$, and RRAR (using ground-truth relevant regions for validation).

\noindent\textbf{Key Finding:} Attention heads exhibit a linear relationship between $R_\text{img}$ and $H_\text{img}$:
$
R_\text{img} = k \cdot H_\text{img} + b
$ ( as shown in Fig.~\ref{fig:SelectHeadsResult}).
Critically, heads that achieve high RRAR---indicating they focus on question-relevant regions---consistently lie in the \textbf{upper-left region} of the $R_\text{img}$-$H_\text{img}$ space.


\noindent\textbf{Cross-model consistency:} Table~\ref{tab:regression_mean_std} shows regression statistics across 5 models. Despite architectural differences, all models exhibit: (1) strong linear correlation (Pearson $r > 0.9$), (2) near-zero intercepts ($|b| < 0.1$). This demonstrates that the $R_\text{img}$-$H_\text{img}$ relationship is a robust property of visual processing in MLLMs, independent of specific architectures.

Importantly, this method requires \textit{no ground-truth annotations}---we identify visual heads purely from attention statistics. This enables its utilization on inference time. 

\noindent\textbf{Key insight:} Effective visual processing requires \textit{both} high attention to images ($R_\text{img}$) \textit{and} spatial concentration (low $H_\text{img}$). High attention alone is insufficient---a head that attends strongly to the entire image but uniformly across all regions fails to localize question-relevant content. Our analysis reveals that \textbf{visual reasoning heads naturally exhibit this dual property}, providing a principled criterion for head selection without supervision.

\begin{table}[htbp]
\centering
\scriptsize
\setlength{\tabcolsep}{1pt}  
\caption{
Linear regression statistics between the image attention ratio ($R_{\text{img}}$) and attention entropy ($H_{\text{img}}$) across multiple MLLMs on the TextVQA dataset.
For each model, we fit the linear relation $R_{\text{img}} = k \cdot H_{\text{img}} + b$ over all visual-question pairs, and report the mean ($\mu$) and standard deviation ($\sigma$) of the slope ($k$), intercept ($b$), and Pearson correlation coefficient ($r$).
A strong positive correlation across models indicates a consistent linear dependency between visual focus and attention dispersion.
}
\label{tab:regression_mean_std}
\renewcommand{\arraystretch}{1.2}
\begin{tabular}{lccc}
\toprule
\textbf{Model} & \textbf{Slope $k$($\mu\pm\sigma$)} & \textbf{Intercept $b$($\mu\pm\sigma$) } & \textbf{Pearson $r$($\mu\pm\sigma$) } \\
\midrule
LLaVA-1.5-7B & $0.167 \pm 0.002$ & $-0.002 \pm 0.003$ & $\textbf{0.991} \pm 0.005$ \\
Ovis2.5-2B & $0.220 \pm 0.015$ & $-0.021 \pm 0.007$ & $\textbf{0.923} \pm 0.036$ \\
Gemma-3-4B & $0.828 \pm 0.042$ & $-0.027 \pm 0.015$ & $\textbf{0.929} \pm 0.020$ \\
Qwen2.5-VL-7B & $0.215 \pm 0.015$ & $-0.028 \pm 0.008$ & $\textbf{0.950} \pm 0.038$ \\
Qwen2.5-VL-3B & $0.213 \pm 0.011$ & $-0.024 \pm 0.008$ & $\textbf{0.951} \pm 0.033$ \\
\bottomrule
\end{tabular}
\end{table}

%% file: sec/5_attention_reweighting.tex
\section{Improving Multimodal Reasoning }
\label{sec:method}
To mitigate the perceptual degradation issue, we utilize our key findings and propose a training-free framework—VRGA.
Our \textbf{Visual Region-Guided Attention (VRGA)} framework, consists of two main components:
(\textit{i}) localization of question-relevant visual regions and 
(\textit{ii}) attention reweighting for response generation.
The overall pipeline is illustrated in Fig.~\ref{fig:vra_overview}.

\subsection{Localization of Question-Relevant Regions}
\label{sec:localization}
To localize question-relevant visual regions without ground-truth annotations, we first identify attention heads that successfully capture meaningful visual information, then mitigate noise in their aggregated attention maps caused by irrelevant tokens, and finally construct a refined attention map from which high-attention tokens are selected as question-relevant regions.

\textbf{Adaptive Head Selection based on Attention-Entropy Dynamics:~}Based on  Sec.~\ref{sec:visualHead}, we propose the \textbf{Entropy--Focus Criterion (EFC)} to dynamically select visual heads that attend to question-relevant visual regions. 
We define the \textit{Entropy--Focus Ratio (EFR)} as
$
\text{EFR} = \frac{H_\text{img}}{R_\text{img}},
$
which captures the trade-off between attention dispersion and image focus.
Lower $\text{EFR}$ indicates more concentrated, image-grounded attention. Combined with $R_{\text{img}}$, these two metrics evaluate each head from three aspects: (1) preference for image over text tokens, (2) coverage of global visual information, and (3) spatial concentration within the image. This allows us to select heads satisfying all three criteria, denoted as $\mathcal{H}_{\text{v}}$, the \textbf{vision-focused heads} that attend to question-relevant visual regions. As shown in Fig.~\ref{fig:SelectHeadsResult} (right), combining $\text{EFR}$ and $R_{\text{img}}$ effectively identifies heads that correctly localize relevant visual regions.

\begin{figure*}[htbp]  
    \centering
    \includegraphics[width=1\linewidth]{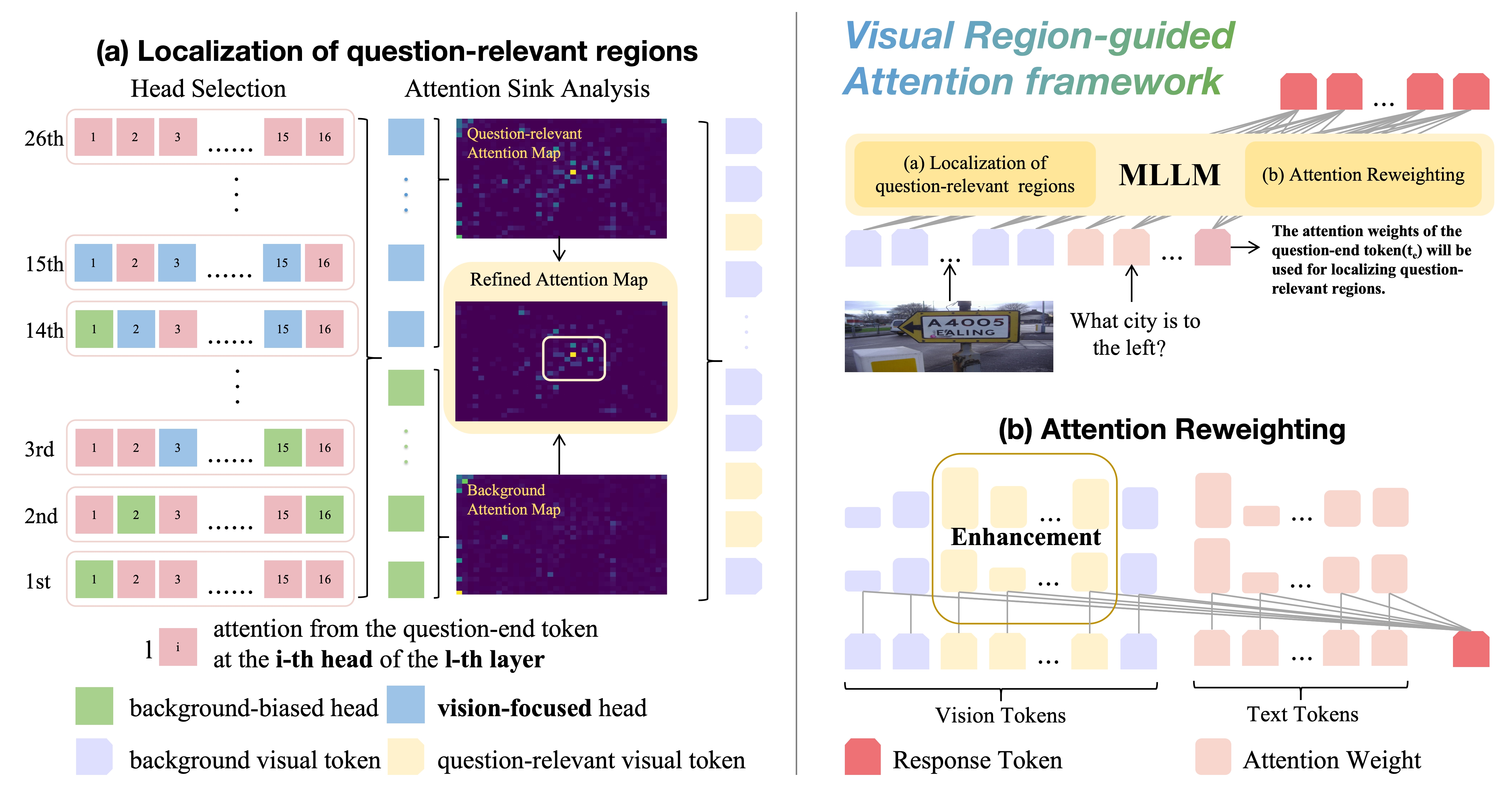}
    \caption{
Overview of the Visual Region-guided Attention (VRGA) framework. Our method enhances the visual grounding capability of Multimodal Large Language Models (MLLMs) without additional training.
(a) \textbf{Localization of Question-relevant Regions:} By integrating the attention maps from vision-focused heads $\mathcal{H}_v$ and background-biased heads $\mathcal{H}_b$, and conducting attention sink analysis, we obtain a clean, refined attention map that facilitates the localization of tokens related to the question.
(b) \textbf{Attention Reweighting:} During the generation phase, attention to tokens within the localized regions is amplified, steering the model to focus on relevant visual content.
}
    \label{fig:vra_overview}
  
\end{figure*}

\textbf{Reducing Noise in Aggregated Attention Maps:~}
\label{sec:attention_sink}
As shown in Fig.~\ref{fig:vra_overview}, even the 
\textbf{question-relevant attention map} 
(aggregated over vision-focused heads) often highlights irrelevant regions, such as tokens in the top-left corner of the image. 
We observe that, for the same image, many heads tend to \textit{sink} to the same non-informative tokens. 

In early layers, the model mainly processes the instruction and has not yet started detailed visual understanding. 
Therefore, we identify a set of heads with low image attention and spatially dispersed patterns in these layers, denoted as $\mathcal{H}_{\text{b}}$, which represent \textbf{background visual heads} that primarily capture low-level or non-informative patterns. 
By analyzing these \textbf{background visual heads}, we can identify which background regions consistently attract attention across heads for the same input, revealing the sources of visual noise.

\textbf{Refined attention map construction:~}
As illustrated in Figure~\ref{fig:vra_overview}, the two types of heads selected by our method tend to focus on the same background tokens. 
To mitigate this \textit{attention sink} effect, we construct refined attention maps as follows. 
For each input sample, we first aggregate the attention maps from all vision-focused heads and normalize them within the image region. 
We then subtract the averaged background-head attention map to suppress sink-induced bias:

\begin{equation}
    \mathbf{A}_{\text{refined}} = 
    \text{Norm}\Bigg(
    \frac{1}{|\mathcal{H}_v|}\sum_{h \in \mathcal{H}_v} \mathbf{A}_h
    - 
    \lambda \cdot 
    \frac{1}{|\mathcal{H}_b|}\sum_{h \in \mathcal{H}_b} \mathbf{A}_h
    \Bigg),
\end{equation}

where $\mathbf{A}_h$ denotes the attention map of head $h$,  
$\mathcal{H}_v$ and $\mathcal{H}_b$ represent the sets of vision-focused and background visual heads, respectively,  
$\lambda$ is a scalar controlling the strength of background subtraction, and  
$\text{Norm}(\cdot)$ normalizes the resulting map within the image region.  
This procedure yields $\mathbf{A}_{\text{refined}}$, which effectively highlights regions with high semantic relevance to the question.

Visual tokens with attention values exceeding a threshold $\tau$ in $\mathbf{A}_{\text{refined}}$ are selected as \textbf{question-relevant visual tokens}:

\begin{equation}
    \mathcal{T}_q = \{ i \mid \mathbf{A}_{\text{refined}}(i) > \tau \},
\end{equation}

where $\tau$ is a predefined threshold controlling token selection.  
These tokens form the basis for the subsequent attention reweighting step.


\begin{table}[t]
\centering
\scriptsize
\setlength{\tabcolsep}{4pt}
\renewcommand{\arraystretch}{1.1}
\caption{
Accuracy (\%) comparison under different head-masking strategies on the \textbf{TextVQA} benchmark.
\textbf{EFR-Guided Masking (ours)} leads to the largest accuracy drop, indicating that the selected heads are most critical for visual reasoning.
}
\label{tab:mask_results}

\begin{tabular}{lcccc}
\toprule
\textbf{Model} & \textbf{Baseline} & \textbf{Random} & \textbf{Low-Visual} & \textbf{EFR-Guided (Ours)} \\
\midrule
Qwen2.5-VL-3B      & 87.64 & 83.38 & 87.02 & \textbf{24.31} \\
Qwen2.5-VL-7B      & 86.88 & 86.95 & 87.50 & \textbf{40.52} \\
Qwen3-VL-30B   & 90.44 & 88.80 & 90.31 & \textbf{42.27} \\
\bottomrule
\end{tabular}

\end{table}

\begin{table*}[tb]
\centering
\footnotesize
\setlength{\tabcolsep}{8pt}
\caption{
Performance comparison on \textbf{HALO}, \textbf{HallusionBench} and \textbf{MMStar}. 
Each column reports the comprehensive score ($S$), accuracy ($ACC\%$), and irrelevance degree ($I$). 
Large Language ModelGA consistently improves grounding quality and reduces irrelevant reasoning.
}
\label{tab:Large Language Modelga_two_benchmarks}
\begin{tabular}{lccccccccc}
\toprule

\multirow{2}{*}{\textbf{Model}}
& \multicolumn{3}{c}{\textbf{HaloQuest}} 
& \multicolumn{3}{c}{\textbf{HallusionBench}}
& \multicolumn{3}{c}{\textbf{MMStar}}\\

\cmidrule(lr){2-4} \cmidrule(lr){5-7} \cmidrule(lr){8-10}
 & \textbf{$ACC\%$ (↑)} & \textbf{$S$ (↑)}  & \textbf{$I$ (↓)} 
 & \textbf{$ACC\%$ (↑)} & \textbf{$S$ (↑)}  & \textbf{$I$ (↓)}
 & \textbf{$ACC\%$ (↑)} & \textbf{$S$ (↑)}  & \textbf{$I$ (↓)}
 \\
\midrule
Qwen3-VL-30B 
& \textbf{69.20} & 0.454  & 0.762 
& 68.55 & 0.507  & \textbf{0.611}
& 66.1 & 0.521  &\textbf{0.542} \\
Qwen3-VL-30B+VRGA(ours)   
& 68.05 & \textbf{0.465}  & \textbf{0.744} 
& \textbf{69.40} & \textbf{0.510}  & 0.614
& \textbf{67.1} & \textbf{0.522}  & 0.543\\

\midrule
Qwen2.5-VL-3B 
& 58.87 & 0.445  & 0.601 
& 53.40 & 0.442  & 0.445
& 49.80 & 0.436  & \textbf{0.382}\\
Qwen2.5-VL-3B+CCOT 
& 41.06  & / & / 
& 53.21 & /  & /
& 46.27 & /  & /\\
Qwen2.5-VL-3B+VRGA(ours)       
& \textbf{59.03}   & \textbf{0.488}  & \textbf{0.405} & \textbf{55.40} & \textbf{0.459}  & \textbf{0.433}
& \textbf{50.93} & \textbf{0.441}  & 0.383\\
\midrule
Qwen2.5-VL-7B 
& 66.67 & 0.502  & 0.595 
& 54.20 & 0.427  & \textbf{0.547}
& 48.53 & \textbf{0.389}  & \textbf{0.557}\\
Qwen2.5-VL-7B+CCOT 
& 51.08 & /  & / 
& 44.83 & /  & /
& / & /  & /\\
Qwen2.5-VL-7B+VRGA(ours)       
& \textbf{73.96} & \textbf{0.549}  & \textbf{0.578} 
& \textbf{55.9} & \textbf{0.444}  & 0.549 
& \textbf{48.73} & 0.388  & 0.569\\

\midrule
Qwen2-VL-7B 
& 53.81 & \textbf{0.406}  & 0.622
& 55.10 & 0.425  & 0.583
& \textbf{52.37} & \textbf{0.419}  & \textbf{0.499}\\

Qwen2-VL-7B+CCOT 
& 29.30 & /  & /
& 51.21 & /  & /
& 50.06 & /  & /\\

Qwen2-VL-7B+ICOT 
& 47.85 & 0.367  & 0.638
& 48.37 & 0.393  & 0.555
& 42.27 & 0.358  & 0.523\\

Qwen2-VL-7B+VRGA(ours) 
& \textbf{54.47} & 0.404  & \textbf{0.608}
& \textbf{57.10} & \textbf{0.438}  & \textbf{0.577}
& 51.07 & 0.405  & 0.513\\

\bottomrule
\end{tabular}
\end{table*}

\subsection{Attention Reweighting for Response Generation}
\label{sec:reweighting}
In this subsection, we aim to enhance the model's attention to question-relevant visual regions during response generation, 
focusing exclusively on \textit{vision-focused heads}. 
Since not all heads are responsible for visual processing, applying attention adjustments to other heads could interfere 
with their original functionality and introduce noise. 
To achieve this, our approach consists of two stages: first, we verify that the selected heads are indeed responsible for visual processing via masking experiments; 
second, we perform attention reweighting on these verified heads to emphasize the question-relevant regions identified in the first stage.

\paragraph{Confirming the Role of Heads.}  To validate that our dynamically selected heads are indeed responsible for visual processing, we conduct \textbf{head-masking experiments} on the \textit{TextVQA} dataset. We zero out their attention to visual tokens ($v_\text{start}$ to $v_\text{end}$) during reasoning:
\begin{equation}
    A_{l,h}[v_\text{start}:v_\text{end}] = 0,
\end{equation}
where $A_{l,h}$ is the attention map of layer $l$ and head $h$.  
A significant drop in accuracy after masking indicates these heads are essential for integrating visual evidence.

To evaluate the effectiveness of our proposed head selection strategy, we compare it against two baseline masking approaches. All methods mask $k$ attention heads per layer during the reasoning stage, where $k=5$ for models with 16 heads per layer (e.g., Qwen2.5-VL-3B, Qwen2.5-VL-7B) and $k=10$ for models with 32 heads per layer (e.g., Qwen3-VL-30B).

\begin{itemize}
    \item \textbf{Random Masking:} Randomly masks $k$ heads per layer.  
    \item \textbf{Low-Visual Masking:} Masks the $k$ heads with lowest image attention ratio $R_\text{img}$.  
    \item \textbf{EFR-Guided Masking (ours):} Selects heads with high visual focus and attention. First, filter the top 50\% by $R_\text{img}$, then pick the $k$ heads with lowest $\text{EFR}$.  
\end{itemize}

The \textbf{Baseline} is the unmodified model using all tokens without masking.
Table~\ref{tab:mask_results} shows that masking the visual heads selected by our method causes a significant accuracy drop compared to Random or Low-Visual Masking.
This confirms that our selected heads are essential for visual reasoning, encoding crucial visual evidence, while other heads mainly handle linguistic or contextual information.

\paragraph{Attention Reweighting.}To guide the model to attend to question-relevant visual regions during reasoning, 
we aim to enhance the attention weights corresponding to these regions. 

We reweight the attention distribution of vision-focused heads when generating the response:
\begin{equation}
    \tilde{\mathbf{A}}_h(i) =
    \begin{cases}
        (1+\gamma)\mathbf{A}_h(i), & \text{if } i \in \mathcal{T}_q, \\
        \mathbf{A}_h(i), & \text{otherwise},
    \end{cases}
\end{equation}
where $\gamma$ controls the intensity of enhancement, and $\tilde{\mathbf{A}}_h$ is then normalized to preserve the total attention mass.

This reweighting effectively guides the model to emphasize the question-relevant regions identified in the first stage, 
leading to more grounded and less hallucinatory responses. Empirically, this simple yet effective reweighting significantly improves visual grounding and reasoning consistency across multiple benchmarks.

%% file: sec/6_experiments_results.tex
\section{Experiments and Results}
\label{sec:results}
\textbf{Evaluation Setup:} We evaluate our \textbf{Visual Region-Guided Attention (VRGA)} framework on three challenging VQA benchmarks:  MMStar~\citep{mmstar}, Hallusion Bench~\cite{guan2024hallusionbench}, and HaloQuest~\citep{wang2024haloquest}. These datasets are carefully chosen to test \textbf{true visual reasoning} capabilities, rather than simple perception or language-based guessing:  

VRGA enhances visual reasoning by guiding models to focus on question-relevant regions, improving the use of visual information.  
To ensure this guidance is effective, we select reasoning-capable models from the Qwen series and test different versions and sizes to evaluate generality and scalability.  
Experiments are conducted with four base models: \textbf{Qwen3-VL-30B}, \textbf{Qwen2.5-VL-3B}, \textbf{Qwen2.5-VL-7B}, and \textbf{Qwen2-VL-7B}, comparing each baseline with its \textit{VRGA}-enhanced version.  
We also compare with chain-of-thought methods CCoT~\cite{ccot} and ICoT~\cite{icot}, which improve reasoning via longer or structured reasoning chains.  

\textbf{Evaluation Metrics:} To jointly assess answer quality and reasoning faithfulness, we introduce three complementary metrics:
(1) \textbf{Binary Correctness ($A$)} $\in \{0, 1\}$, measuring the semantic correctness of the response;
(2) \textbf{Irrelevance Degree ($I$)} $\in [0, 1]$, quantifying the degree of off-topic or irrelevant content (0 indicates no irrelevant content);
and (3) a \textbf{Comprehensive Score} computed as:
\[
S = A \times (1 - \alpha I),
\]
where $\alpha$ is a penalty coefficient controlling the influence of irrelevant information.

\textbf{Results:} As shown in Table~\ref{tab:Large Language Modelga_two_benchmarks}, VRGA consistently improves performance on the evaluated benchmarks and model scales. For instance, on HaloQuest, the comprehensive score for Qwen2.5-VL-3B increases from \textbf{0.445} to \textbf{0.488}, mainly due to a reduction in irrelevance ($I$: 0.601 $\rightarrow$ 0.405) while maintaining accuracy, and for Qwen2.5-VL-7B the score rises from \textbf{0.502} to \textbf{0.549} with higher answer correctness ($A$: 0.667 $\rightarrow$ 0.740) and reduced topic drift. A similar trend is observed on HallusionBench and MMStar, where VRGA consistently reduces irrelevant reasoning and improves the comprehensive score, even if absolute gains in accuracy are modest. These results demonstrate that guiding the model to focus on question-relevant visual regions effectively enhances reasoning precision, mitigates hallucinations, and bridges the reasoning–perception gap in multimodal large language models.

%% file: sec/7_conclusions.tex
\section{Conclusions}

In this work, we investigate the causes of reasoning degradation in multimodal LLMs for visual question answering. We find that chain-of-thought prompts often induce attention dispersion, causing models to lose focus on question-relevant regions despite largely correct perceptual descriptions. Layer- and head-level analyses reveal that effective visual processing correlates with spatially concentrated attention, explaining CoT-induced performance drops. 
Based on these insights, we propose \textbf{Visual Region-guided Attention (VRGA)}, which dynamically identifies visual heads and enhances their attention to question-relevant regions at inference time, restoring perceptual grounding without additional training. Extensive experiments across multiple benchmarks show that guiding attention improves reasoning accuracy and robustness. 
These findings also suggest a promising future direction: leveraging reinforcement learning to train models to perform visually grounded reasoning beyond post-hoc attention reweighting.